\documentclass{elsarticle}
\usepackage{hyperref}
\journal{Neurocomputing}

\usepackage{amsmath,amsfonts}
\usepackage{graphicx,subfig}
\begin{document}

\begin{frontmatter}

\title{Multifidelity Bayesian Optimization for Binomial Output}
% \tnotetext[mytitlenote]{Fully documented templates are available in the elsarticle package on \href{http://www.ctan.org/tex-archive/macros/latex/contrib/elsarticle}{CTAN}.}

\author[mymainaddress,hseaddress]{Leonid Matyushin}

\author[mymainaddress,mysecondaryaddress]{Alexey Zaytsev\corref{mycorrespondingauthor}}
\cortext[mycorrespondingauthor]{Corresponding author}
\ead{a.zaytsev@skoltech.ru}

\author[mymainaddress]{Oleg Alenkin}

\author[hseaddress]{Andrey Ustuzhanin}

\address[mymainaddress]{Skoltech, Moscow}
\address[mysecondaryaddress]{IITP RAS, Moscow}
\address[hseaddress]{National Research University Higher School of Economics, Moscow}

\begin{abstract}
The key idea of Bayesian optimization is replacing an expensive target function with a cheap surrogate model. By selection of an acquisition function for Bayesian optimization we trade off between exploration and exploitation. The acquisition function typically depends on mean and variance of the surrogate model at a given point.
The most common Gaussian process-based surrogate model assumes that the target with fixed parameters is a realization of a Gaussian process. However, often the target function doesn't satisfy this approximation. Here we consider target functions that come from the binomial distribution with the parameter that depends on inputs.
Typically we can vary how many Bernoulli samples we obtain during each evaluation. 
We propose a general Gaussian process model that takes into account Bernoulli outputs. To make things work we consider a simple acquisition function based on Expected Improvement and a heuristic strategy to choose the number of samples at each point thus taking into account precision of the obtained output.
\end{abstract}

\begin{keyword}
Bayesian optimization \sep Gaussian processes \sep Binomial distribution \sep Multifidelity
\end{keyword}

\end{frontmatter}

% \linenumbers

\section{Introduction}
\label{introduction}

Bayesian optimization (BO) is a powerful class of optimization methods that allows optimization of black-box non-deterministic functions. In vanilla approach we assume that this function is a deterministic function plus Gaussian noise and then obtain analytical treatment of the problem of evaluation of posterior mean and variance.
Using mean and variance we can evaluate most of the acquisition functions used for selection of point for evaluation at the next step of optimization~\cite{shahriari2016taking}. 

This assumption about the target function is sometimes inadequate.
For example, for binomially distributed observations vanilla BO sometimes struggles to find a minimum, as the model is wrong. 
The binomially distributed observations often occur in high energy physics, e.g. the spectrometer tracker optimization \cite{van2015simulation} and the muon shield optimization \cite{baranov2017optimising}. 
In both examples, target functions are Monte-Carlo simulations of real experiments with main reasons of randomness are quantum effects.

These two examples share many properties.
The target functions are expensive to evaluate. 
It took hours to get results using a modern cluster.
The target functions have discrete distribution, so they are not Gaussian-noised deterministic functions. 
For example, the last one is naturally Binomial distributed. 
It is possible to choose the complexity of simulation determined by the number of simulated particles. 
High-fidelity simulations are accurate but expensive. 
Low-fidelity simulations are cheaper but less accurate.  

Thus we need approach that able to deal with this kind of problems.
To create such approach we need to propose a correct model based on Generalized Gaussian Process regression, then construct an acquisition function.
Also we need to clarify if we can improve our models using availability of multifidelity evaluations.

The paper is organized as follows. In Section \ref{preliminary} we give a preliminary information about Bayesian Optimization (BO) and Gaussian processes (GP). Section \ref{proposed approach} is devoted to our modification of GP model construction and BO for binomial output. 
In section \ref{experiments} we investigate usefulness of the proposed approach and examine peculiarities and possible applications.
We use artificial functions in our numerical experiments.

\section{Related Works}
\label{related works}

Area of application of Bayesian optimization known in different areas under different names are quite wide. 
A recent overview of Bayesian optimization is provided by authors in~\cite{shahriari2016taking},
see this article and references in it. 
Below we cover some issues related to our specific applications. 

We start of range of applications where the output is binomial.
See e.g. problems of hyperparameter tuning or AutoML: In work \cite{domhan2015speeding} authors propose an early stopping criterion combined with modification of EI acquisition function in which evaluation of a configuration is stopped if predicted performance is worse than the current best configuration. Bayesian optimization was used for tuning of hyperparameters for Alpha Go~\cite{chen2018bayesian} as well as for other deep learning based systems~\cite{klein2016fast}.
Also see \cite{van2015simulation} and \cite{baranov2017optimising}
for high energy physics.

As mentioned in chapter \ref{introduction} values of a black box could have Binomial distribution. It means that the exact Bayesian inference fails, since the likelihood is not Gaussian. 
The same problem arises when you try to adapt the Gaussian processes for the task of classification \cite{nickisch2008approximations} or robust regression with Laplace or Cauchy likelihood \cite{opper2009variational}. 

To use these models one can approximate non-Gaussian posterior by Gaussian distribution. Many approaches are used in this area, to name a few~\cite{bishop2006pattern}: Markov-chain Monte Carlo \cite{nickisch2008approximations}, Laplace approximation \cite{williams2006gaussian}, mean field variational inference~\cite{blei2017variational}, and expectation propagation \cite{minka2001expectation}. 
GP models like GP classifier, GP counter or GP regression use different observations likelihoods: Bernoulli, Poisson, Gaussian, Binomial and etc. All these distributions are samples of exponential family. Aim of the work \cite{shang2013approximate} is to show how to create a framework unifying all existing GP models and making easier creating of new ones using distribution from exponential family. 

Common Bayesian optimization approaches assume single-fidelity simulations. But in some cases it is possible to use cheaper calculation of the same objective with lower fidelity. For example, in such fields as aerodynamics, hyperparameter tuning and industrial design there is an opportunity to use simulations with different fidelities \cite{forrester2007multi,huang2006sequential,klein2015towards,zaytsev2016variable,zaytsev2016reliable}. 
Methods considered above consider specific problems. More general approach MF-GP-UCB \cite{kandasamy2016gaussian} considers a finite number of approximations and assumes that there exist an upper bound for the difference between high and low-fidelity. This model doesn’t allow sharing information between fidelities, and each model is treated independently. 
There exist a generalization of this approach to the continuous fidelity case \cite{kandasamy2017multi}. 

% TODO Bayesian optimization for nonGaussian cases: search for articles

\section{Overview}
\label{preliminary}

\subsection{Bayesian optimization}

We want to minimize $f(\mathbf{x})$ function. 
Suppose that it is impossible to evaluate it directly.
In classic case for Gaussian process regression we observe $\mathcal{N}(f(\mathbf{x}), \sigma^2)$.
In this paper we consider observations from the $\mathrm{Bin}(N, f(\mathbf{x}))$, where $N$ is a number of evaluations at each point.
We can represent a vanilla BO as the following iterative scheme: 

\begin{enumerate}
\item Train a regression model that approximates target function via GP. Now we can evaluate an acquisition function $C_*(\mathbf{x})$ using the regression model.
\item Obtain the point that maximizes the acquisition function
\[
\mathbf{x}_{i + 1} = \mathrm{argmax} C_*(\mathbf{x}),
\]
\item Evaluate 
\[
y_{i + 1} = \mathcal{N}(f(\mathbf{x}_{i + 1}), \sigma^2 
\]  
\item Update the available sample 
\[ 
D = (\mathbf{X}, y) = (\mathbf{X} \cup \mathbf{x}_{i + 1}, y \cup y_{i + 1})
\]
\end{enumerate}

Now let us consider each step in more details.

\subsection{Regression Model}

Gaussian process regression is a popular approach for the construction of nonlinear regression models~\cite{williams2006gaussian} with uncertainty estimates required to perform Bayesian optimization.

Gaussian process on the $\mathbf{R}$ is fully specified by mean $m(\mathbf{x})$ and covariance functions $k(\mathbf{x}, \mathbf{x}')$. 
Following Bayesian ideology on k-th step we put the following prior distribution over $f_i$'s, where $f_i = f(\mathbf{x}_i)$:
\[
p(\mathbf{f}|\mathbf{X}) = N(\mathbf{\mu}, \mathbf{\Sigma}),
\]
where $\mathbf{\mu} = (m(x_i))_{i=1}^k$ and $\mathbf{\Sigma} = (k(x_i, x_j))_{i,j=1}^{k,k}$. 
The typical way is to set $m \equiv 0$, $k(x, x') = \sigma_f^2exp(-\frac{\|x - x' \|_2^2}{2\theta^2})+\sigma_n^2\delta_{i,j}$. 

To perform a Bayesian inference we need to choose likelihood. Classical approach suggest us to work with a Gaussian likelihood, since in this case prior and likelihood are conjugate and it is possible to find an exact expression for a posterior~\cite{burnaev2011modeling}.

\[
p(\mathbf{f}|\mathbf{X},\mathbf{y}) = \frac{p(\mathbf{y}|\mathbf{f})p(\mathbf{f}|\mathbf{X})}{\int p(\mathbf{y}|\mathbf{f})p(\mathbf{f}|\mathbf{X}) d\mathbf{f}} 
\label{eq:1}
\]

This is a distribution over already visited points. To make a prediction we need to compute the following integral. 
In case of a Gaussian likelihood we have two conjugate distribution inside of this integral, so result could be found analytically. 

\[
p(\mathbf{f_*}|\mathbf{X},\mathbf{y},\mathbf{x_*}) = \int p(\mathbf{f_*}|\mathbf{f},\mathbf{X},\mathbf{x_*}) p(\mathbf{f}|\mathbf{X},\mathbf{y})d\mathbf{f}
\label{eq:2}
\]

Where $p(\mathbf{f_*}|\mathbf{f},\mathbf{X},\mathbf{x_*})$ is a marginal Gaussian, since all $f_i$'s are Gaussian distributed. And the distribution over noised variable is nothing but

\[
p(\mathbf{y_*}|\mathbf{X},\mathbf{y},\mathbf{x_*}) = \int p(\mathbf{y_*}|\mathbf{f_*}) p(\mathbf{f_*}|\mathbf{X},\mathbf{y}, \mathbf{x_*})d\mathbf{f_*}
\label{eq:3}
\]

Where $p(\mathbf{y_*}|\mathbf{f_*})$ is a Gaussian distribution in case of a Gaussian likelihood, so the classical final answer is a Gaussian distribution. 
It follows that the posterior mean $\mu(\mathbf{x_*})$ and the posterior variance $\sigma^2(\mathbf{x_*})$ have the following form:
\begin{align*}
    \mu(\mathbf{x_*}) &= \mathbf{E}[\mathbf{y_*}|\mathbf{X},\mathbf{y},\mathbf{x_*}] = K(\mathbf{x_*}, \mathbf{X})K(\mathbf{X}, \mathbf{X})^{-1}\mathbf{f}, \\
    \sigma^2(\mathbf{x_*}) &= \mathbf{V}[\mathbf{y_*}|\mathbf{X},\mathbf{y},\mathbf{x_*}] = K(\mathbf{x_*},\mathbf{x_*})-K(\mathbf{x_*},\mathbf{X})K(\mathbf{X}, \mathbf{X})^{-1}K(\mathbf{X}, \mathbf{x_*}).
\end{align*}

\subsection{Acquisition function}

At each iteration of Bayesian optimization we select the next point to evaluate the target function. 
There are several approaches. 
One of the most popular choice we adopt here is ``Expected Improvement'' (EI): 
\[
C_{EI}(\mathbf{x}) = \mathbf{E}[\max (0, y_{min}-f(\mathbf{x}))|\mathbf{X},\mathbf{y}],
\label{eq:EI}
\] 
where $y_{min}$ is the minimal value obtained to the current iteration. 
In case of a Gaussian likelihood it was shown that it converges under mild assumption [citation TODO]
and has a closed-form expression:
\[ 
C_{EI}(\mathbf{x}) = (y_{min}-\mu(\mathbf{x}))\mathrm{\Phi}(y_{min}|\mu(\mathbf{x}),\sigma^2(\mathbf{x}))+\sigma^2(\mathbf{x})\mathcal{N}(y_{min}|\mu(\mathbf{x}),\sigma^2(\mathbf{x})),
\]
Where $\mathrm{\Phi}$ is the standard normal CDF. 

Next point is $\mathrm{argmax}$ of $C_{EI}$. 
Expected improvement has many local maxima and regions with almost constant function value.
As we need to evaluate only the posterior mean and variance at each point, it is cheap to have as many evaluation as required,
so most of the global optimization methods can solve this task.

\section{Proposed approach}
\label{proposed approach}

In this paper we consider Binomial Bayesian optimization, so observable variable have Binomial distribution:

\[
y_{i + 1} = \mathrm{Bin}(N, f(\mathbf{x}_{i + 1}))
\]

In this case we use a GGPM setting to perform a Bayesian Inference \cite{shang2013approximate}. In particular, \eqref{eq:1}, \eqref{eq:2} \eqref{eq:3} could not be expressed in a closed form, since in this case there is no conjugate distributions in the corresponding formulas, so one should use approximate methods, such as Laplace approximation \cite{shang2013approximate}. We approximate Expected Improvement \eqref{eq:EI} in Binomial setting via Monte-Carlo.

In practice we specify parameter $N$ for Binomial distribution before evaluation of the black-box. 
Low $N$'s allow us to spend less computational resources, while large $N$'s allows to make a more exact evaluation.

The proposed method considers two different fidelities for simulation $N_{low}$ and $N_{high}$. 
First thing should be done is determining of these fidelities. To distinguish promising point one can run low fidelity simulation at this point that does not cost a lot. Then using obtained information one can make a decision to continue simulation at the same point or to move to the next one proposed by acquisition function.

Decision function $d$ takes as input result of low fidelity simulation at current point $\mathbf{x_*}$, actual surrogate model and some external parameters. After running low fidelity simulation one can calculate posterior distribution of objective function at this point. Observation $y_{low}$ is a realization of Binomial random value with parameters $p=f(\mathbf{x_*})$ and $N=N_{low}$, one can estimate posterior distribution of $f(\mathbf{x_*})$:
$$p(f(\mathbf{x_*})|y_{low}) = \frac{p(y_{low}|f(\mathbf{x}))p(f(\mathbf{x}))}{ p(y_{low})}$$
$y_{low}|f(\mathbf{x})$ is binomial distributed, so we assume that $p(f(\mathbf{x}))$ is Beta distribution with parameters $\alpha=1$, $\beta=1$ (uniform distribution on $[0, 1]$) since in this case we can perform Bayesian inference in a closed form:
$$f(\mathbf{x_*})|y_{low} \sim \mathrm{B}(1 + y_{low}, 1 + N_{low} - y_{low})$$

Let us now compute the probability of improvement of $y_{min}$ at the current point:
$$\mathrm{P}(f(\mathbf{x}) < y_{min}) = \int_{-\infty}^{y_{min}}\mathrm{B}(\tau | 1 + y_{low}, 1 + N_{low} - y_{low})d\tau $$
This probability is the value of Beta CDF at $y_{min}$. 
Now one compare this probability with the threshold $\lambda$ and make a decision to
continue calculation at current point or to spend available budget for evaluations at other points. 
In section \ref{experiments} we explore performance of proposed approach for different choices of~$\lambda$.  

\section{Experiments}
\label{experiments}

\subsection{Methodology of comparison}

We assume that workflow of Bayesian optimization and the only computational expenses are related to evaluations of a black box  $f(x)$. For each function we consider how the minimal seen value (i.e. $\frac{f(x)}{N}$) depends on computational resources.
We suppose that high fidelity is $\frac{N_{high}}{N_{low}}$ times more expensive than low fidelity. 

To compare the approach proposed in Section~\ref{proposed approach} with other methods we perform massive testing of considered algorithms on different optimization problems. 
Each problem is characterized by a target function and the parameter $N$. Optimization method starts from random initial design and performs several steps of optimization of given function. 
For the sake of comparison we rescale optimization results with respect to spent computational resources. As a result, for each problem we got a minimal seen values for a unit of computational resources. We take an average of this values over multiple runs to achieve stability.

\subsection{Metrics}

We use Dolan-More curves as a method to compare results of optimization \cite{dolan2002benchmarking,belyaev2016gtapprox}. 
To define Dolan-More curves for our problem we need to specify a set of \textit{problems} $\mathcal{P}$, a set of \textit{solvers} $\mathcal{S}$, and $t_{p,s}$ --- the measure of success of an approach $s$ on a problem $p$. 
In our case $\mathcal{P}$ was defined in previous paragraph - it is a set of all problems of minimization of described functions for a fixed budget. 

$\mathcal{S}$ is a set that consists of vanilla Gaussian Bayesian optimization, vanilla Binomial Bayesian optimization, and proposed modification of Binomial Bayesian optimization with $\lambda = 0.3$ and $\lambda = 0.5$. 
For all methods we used SLSQP method to maximize Expected Improvement. $t_{p,s}$ is a true value of objective function at the point with minimal observed value at the current step (averaged over multiple runs).

Finally the Dolan-More curve is the following:
\begin{align*}
r_{p,s} &= \frac{t_{p,s}}{\min_{s \in \mathcal{S}}t_{p,s}} \\
rho_{s}(\tau) &= \frac{1}{\#\mathcal{P}} \#\{ p \in \mathcal{P} | r_{p,s} < \tau \}
\end{align*}

The higher is curve the better is corresponding solver. 

\subsection{Generation of samples for evaluation}

Artificial functions of different dimensions are a common choice for benchmarking of test optimization algorithms. 
For our case the objective function should lie in the interval $[0, 1]$. 
So, we rescale the artificial function $\hat{f}(x)$ using minimum $\hat{f}_{\min}$ and maximum $\hat{f}_{\max}$ values of them: 
\[
f(x) = \frac{\hat{f}(x)-\hat{f}_{\min}}{\hat{f}_{\max}-\hat{f}_{\min}}
\]
The values $\hat{f}_{\min}$ and $\hat{f}_{\max}$ are not always known, so we obtained them using numerical optimization.
In the end we have $f(x)$ that lies in the desired interval and can be interpreted as the probability of a success at a particular point. 

We used several popular artificial functions $\hat{f}(x)$: Michalewicz, Rastrigin, Zakharov and Styblinski-Tang functions on hypercubes in $\mathbb{R}^4$, $\mathbb{R}^5$ and $\mathbb{R}^6$. 
Observations of these functions are sampled from Binomial distribution with parameters $f(x)$ and $N$ with $N$ specified manually.

\subsection{Results}

\begin{figure}%
    \centering
    \subfloat[Zakharov]{{\includegraphics[width=6cm]{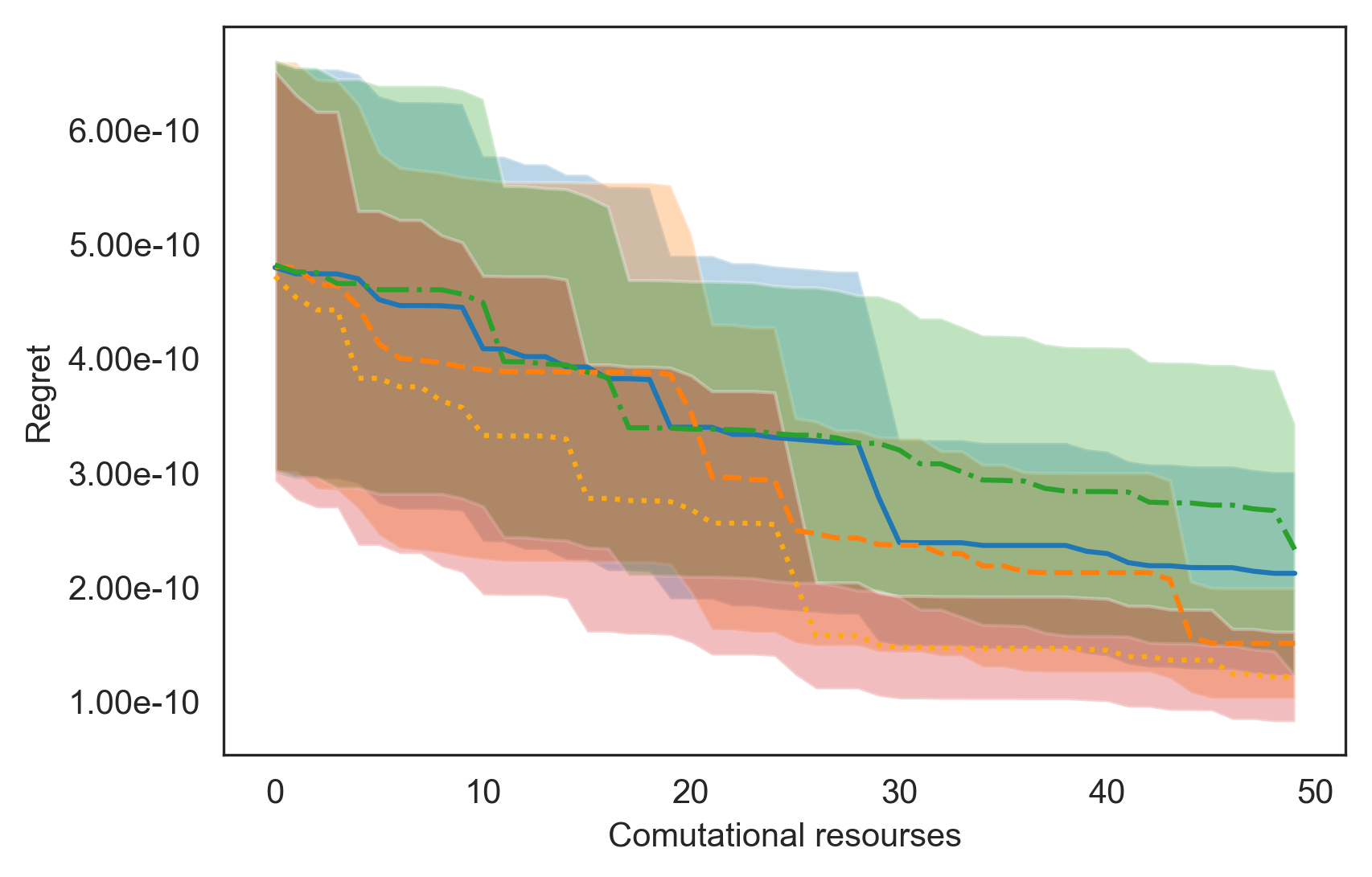} }}
    \subfloat[Styblinski-Tang]{{\includegraphics[width=6cm]{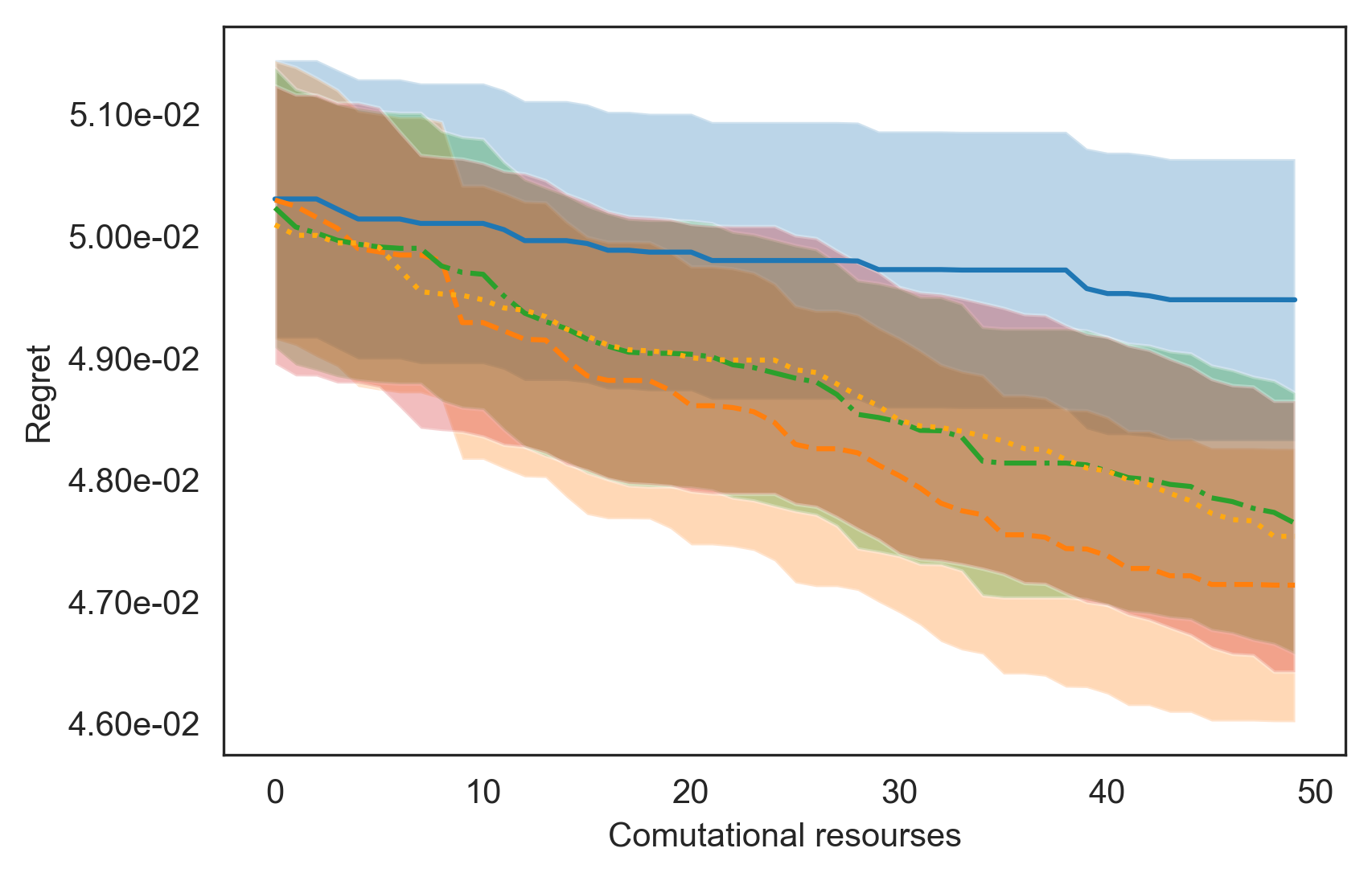} }}%
    \caption{Optimization process for functions on $\mathrm{R}^5$, $N = 70$ with one minimum}%
    \subfloat[Michalewicz]{{\includegraphics[width=6cm]{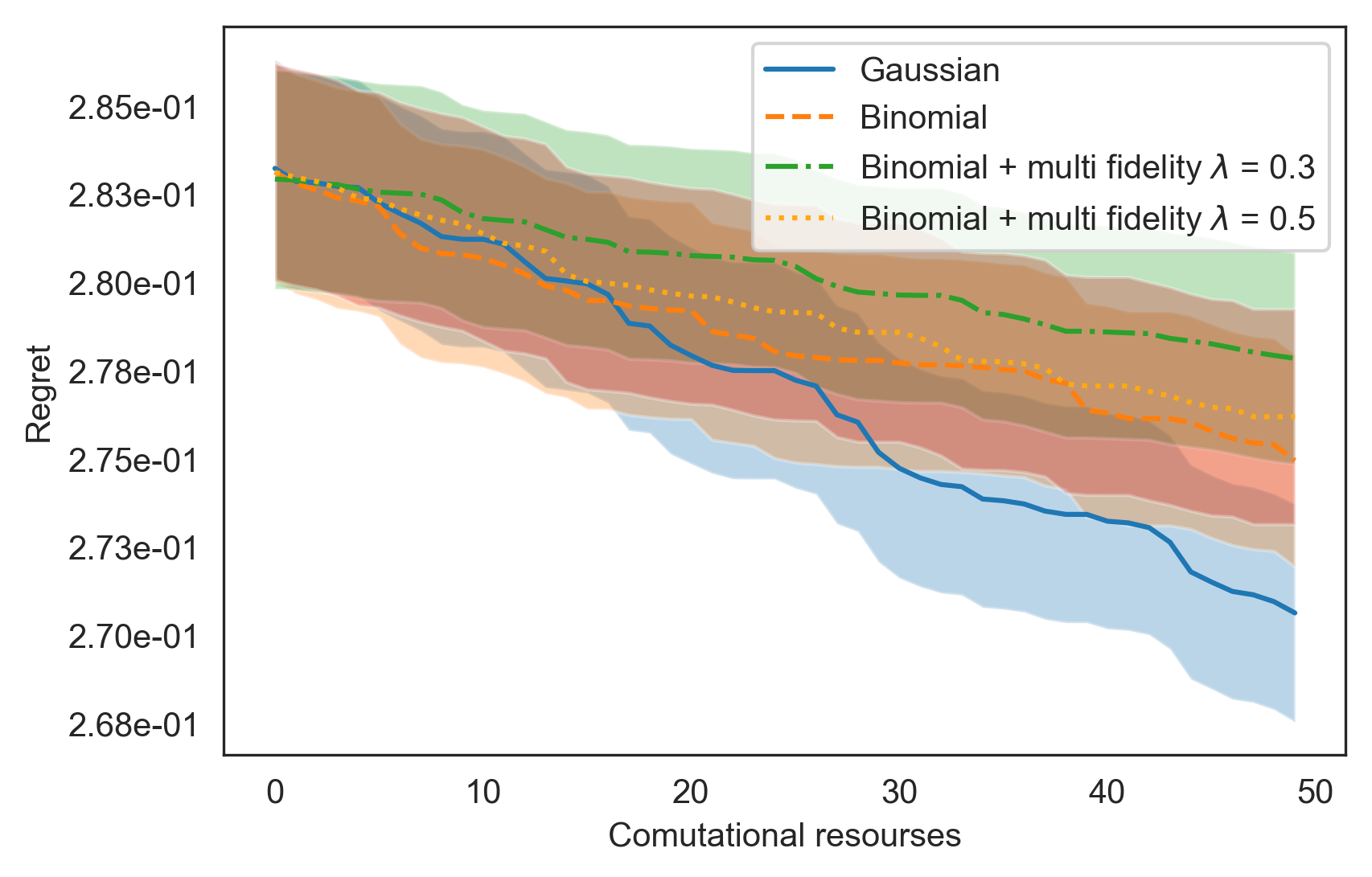} }}%
    \subfloat[Rastrigin]{{\includegraphics[width=6cm]{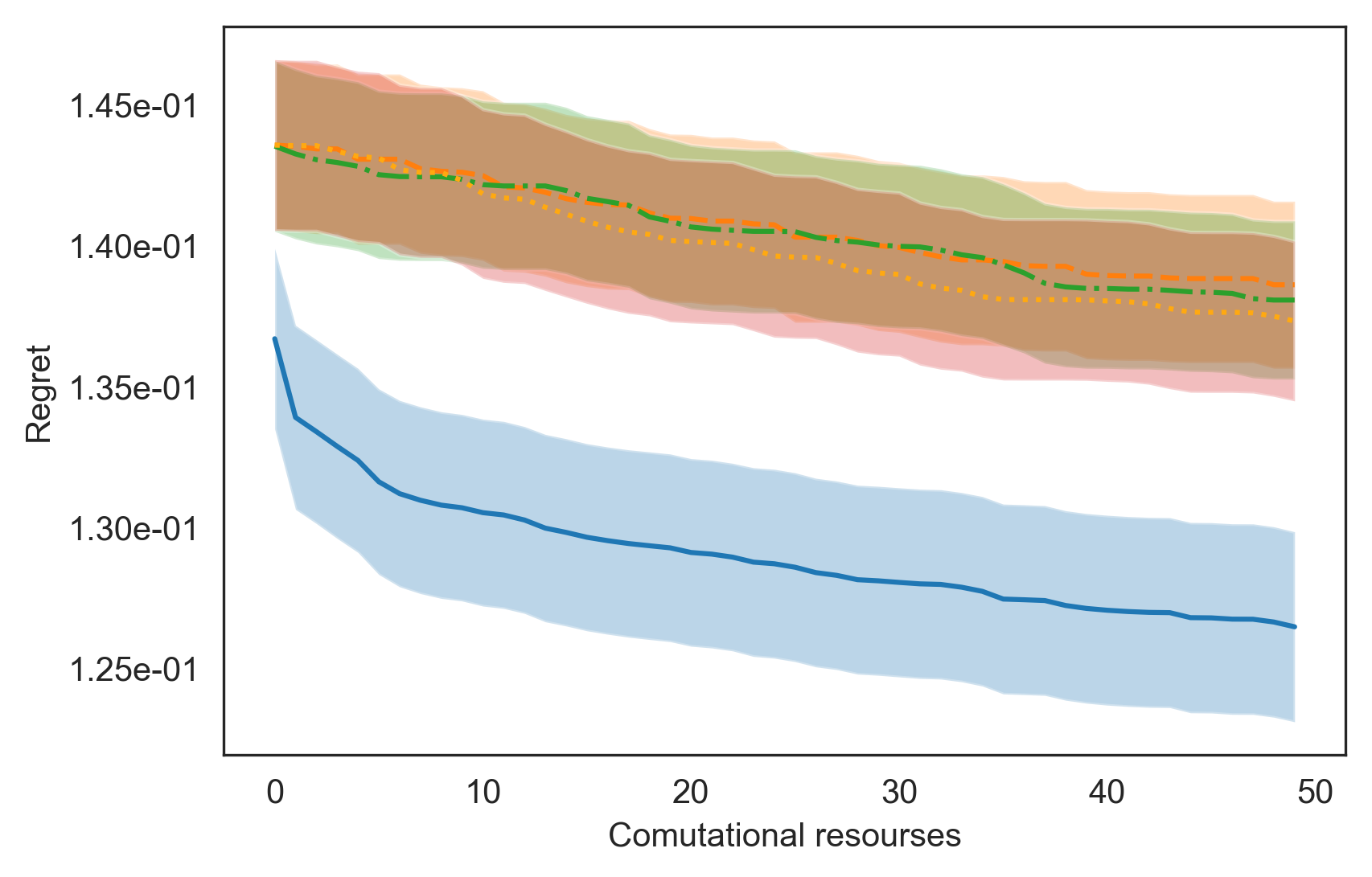} }}%
    \caption{Optimization process for functions on $\mathrm{R}^5$, $N = 70$ with multiple minima. The x axes correspond to the number of devoted computational resources. For Zakharov and Styblinski-Tang functions vanilla approach works worse, while it works better for Michalewicz and Rastrigin functions. The y axes correspond to regret values.}%
    \label{fig:grpahs}%
\end{figure}

Note, that two of considered functions Zakharov and Styblinski-Tang have only one global optimum, while the other two Michalewicz and Rastrigin have multiple extremes.

\begin{figure}%
    \centering
    \subfloat[Functions with a single global minimum ]{{\includegraphics[width=6cm]{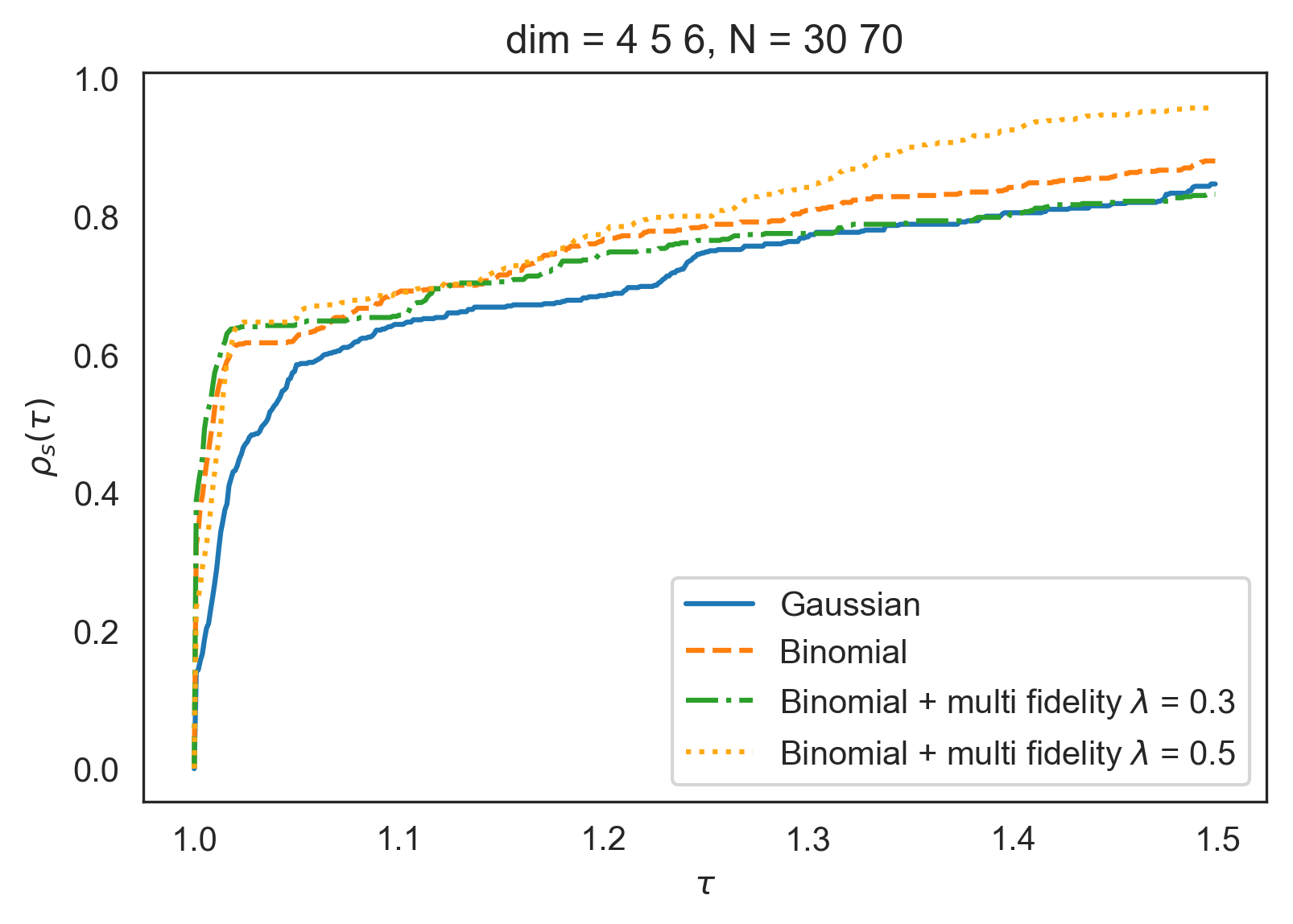} }}%
%    \qquad
    \subfloat[Functions with multiple local minima]{{\includegraphics[width=6cm]{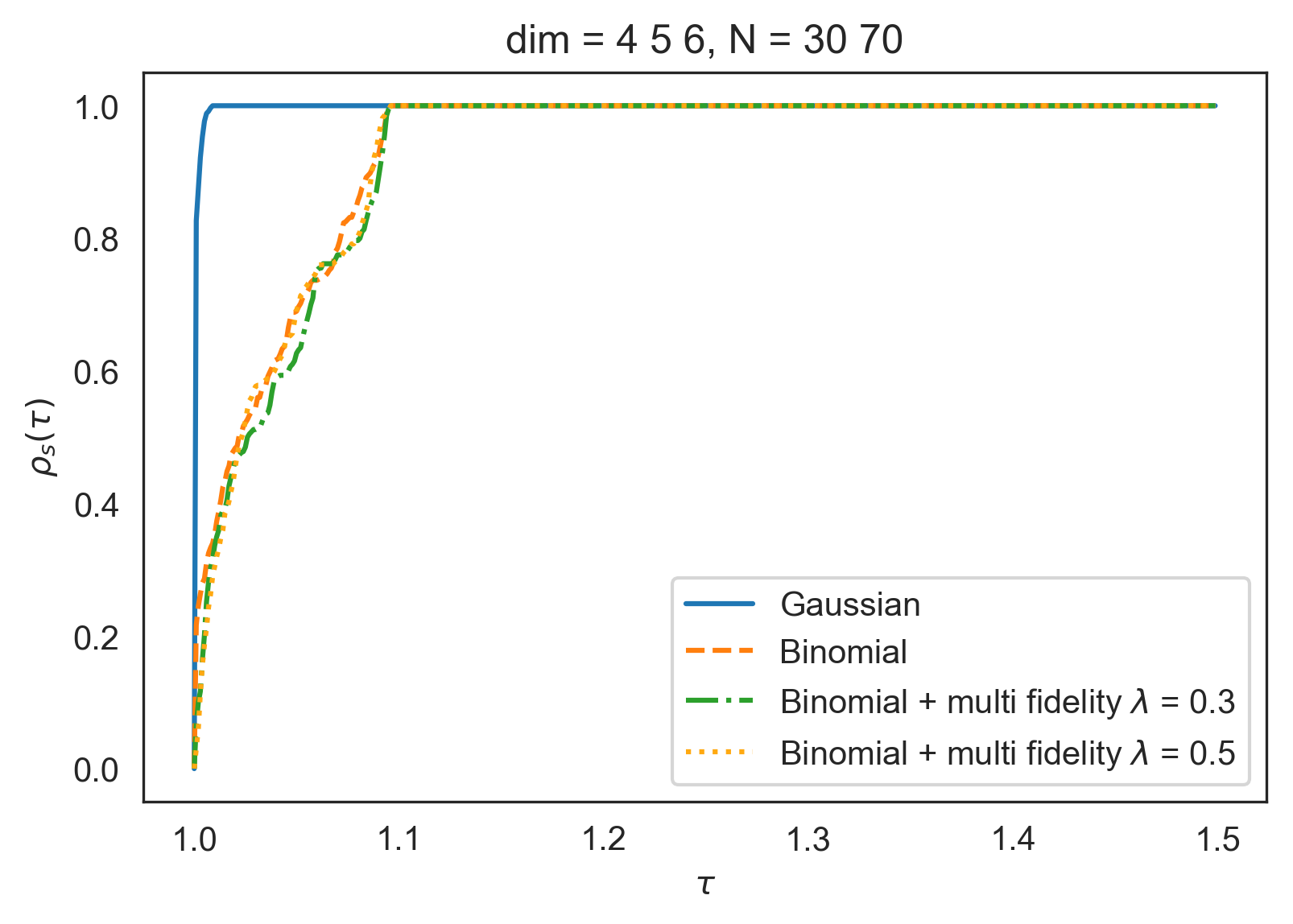} }}%
    \caption{Dolan-More curves for considered algorithms }%
    \label{fig:dolanmore}%
\end{figure}

Figures~\ref{fig:grpahs} depict dynamic of regret w.r.t number of devoted computational resources. 
We see that the behaviour is different for different models used.
 At the figure \ref{fig:dolanmore} we see can see Dolan-More curves for each of these two groups. 
 The difference is even more evident.

We observe that in the single optimum case vanilla Binomial Bayesian optimization outperform vanilla Gaussian Bayesian optimization. Also for $\lambda = 0.5$ Binomial Bayesian outperform all other algorithms. In multiple extremes case situation is different: vanilla Gaussian Bayesian optimization outperforms Binomial algorithms. We conclude, that Binomial optimization approach and, especially, the multifidelity one is a good choice for single optimum functions, while for multiple extreme function vanilla Gaussian Bayesian optimization is better.

\section{Discussion}

In this paper we consider the problem of optimization in non-Gaussian likelihood setting. According to our findings usage of proper surrogate model during Bayesian optimization significantly improve performance. It is important for the case of complex multi-optimum functions. But at the same time this approach has a drawback: using of GGPM instead of usual GP regression for approximation fitting requires more times, approximated inference instead of closed form calculation should be used. As a consequence calculation of acquisition function also becomes intractable problem, in this work we applied Monte-Carlo sampling technique to this problem. But this drawback is not crucial, because BO is usually applied to heavy to evaluate black-box function, so time spent on searching of next point is much lower than time of objective function evaluation. 

We also proposed modifications of the algorithm to work with variable fideltiy evaluations: high cost of simulation and low level of noise, low cost of simulation and high level of noise. 
The proposed heuristics showed weak improvement in comparison with single-fidelity optimization. 
For further improvement one can try to select hyperparameters $N_{high}$ and $N_{low}$ fidelity in a smarter way. 

In this work the ratios $\frac{N_{high}}{N_{low}}$ for conducted experiments were chosen close to 2. But we assume that using of reinforcement learning technique for selection these parameters depending on particular function might led to better performance of optimization. 

\section{Conclusions}

Bayesian optimization is actively used today in different fields. There are several main directions for the development of this approach. Firstly, computation not one but a batch of next points to parallelize evaluation of costly function. Secondly, working with different fidelities. In this work we showed that the usage of suitable surrogate models gives significant improvement of optimization. Moreover we have proposed algorithm suitable for all distributions from the exponential family.

\section{Acknowledgments}
\label{acknowledgments}

The research, presented in Section \ref{experiments} of this paper, was partially supported by the Russian Foundation for Basic Research grants 16-01-00576 A and 16-29-09649 ofi m.

\section*{References}

\bibliography{mybibfile}

\begin{thebibliography}{10}
\expandafter\ifx\csname url\endcsname\relax
  \def\url#1{\texttt{#1}}\fi
\expandafter\ifx\csname urlprefix\endcsname\relax\def\urlprefix{URL }\fi
\expandafter\ifx\csname href\endcsname\relax
  \def\href#1#2{#2} \def\path#1{#1}\fi

\bibitem{shahriari2016taking}
B.~Shahriari, K.~Swersky, Z.~Wang, R.~P. Adams, N.~De~Freitas, Taking the human
  out of the loop: A review of bayesian optimization, Proceedings of the IEEE
  104~(1) (2016) 148--175.

\bibitem{van2015simulation}
E.~Van~Herwijnen, T.~Ruf, M.~Ferro-Luzzi, H.~Dijkstra, Simulation and pattern
  recognition for the ship spectrometer tracker, Tech. rep. (2015).

\bibitem{baranov2017optimising}
A.~Baranov, E.~Burnaev, D.~Derkach, A.~Filatov, N.~Klyuchnikov, O.~Lantwin,
  F.~Ratnikov, A.~Ustyuzhanin, A.~Zaitsev, Optimising the active muon shield
  for the ship experiment at cern, in: Journal of Physics: Conference Series,
  Vol. 934, IOP Publishing, 2017, p. 012050.

\bibitem{domhan2015speeding}
T.~Domhan, J.~T. Springenberg, F.~Hutter, Speeding up automatic hyperparameter
  optimization of deep neural networks by extrapolation of learning curves, in:
  Twenty-Fourth International Joint Conference on Artificial Intelligence,
  2015.

\bibitem{chen2018bayesian}
Y.~Chen, A.~Huang, Z.~Wang, I.~Antonoglou, J.~Schrittwieser, D.~Silver,
  N.~de~Freitas, Bayesian optimization in alphago, arXiv preprint
  arXiv:1812.06855.

\bibitem{klein2016fast}
A.~Klein, S.~Falkner, S.~Bartels, P.~Hennig, F.~Hutter, Fast bayesian
  optimization of machine learning hyperparameters on large datasets, arXiv
  preprint arXiv:1605.07079.

\bibitem{nickisch2008approximations}
H.~Nickisch, C.~E. Rasmussen, Approximations for binary gaussian process
  classification, Journal of Machine Learning Research 9~(Oct) (2008)
  2035--2078.

\bibitem{opper2009variational}
M.~Opper, C.~Archambeau, The variational gaussian approximation revisited,
  Neural computation 21~(3) (2009) 786--792.

\bibitem{bishop2006pattern}
C.~M. Bishop, Pattern recognition and machine learning, springer, 2006.

\bibitem{williams2006gaussian}
C.~K. Williams, C.~E. Rasmussen, Gaussian processes for machine learning,
  Vol.~2, MIT Press Cambridge, MA, 2006.

\bibitem{blei2017variational}
D.~M. Blei, A.~Kucukelbir, J.~D. McAuliffe, Variational inference: A review for
  statisticians, Journal of the American Statistical Association 112~(518)
  (2017) 859--877.

\bibitem{minka2001expectation}
T.~P. Minka, Expectation propagation for approximate bayesian inference, in:
  Proceedings of the Seventeenth conference on Uncertainty in artificial
  intelligence, Morgan Kaufmann Publishers Inc., 2001, pp. 362--369.

\bibitem{shang2013approximate}
L.~Shang, A.~B. Chan, On approximate inference for generalized gaussian process
  models, arXiv preprint arXiv:1311.6371.

\bibitem{forrester2007multi}
A.~I. Forrester, A.~S{\'o}bester, A.~J. Keane, Multi-fidelity optimization via
  surrogate modelling, Proceedings of the royal society a: mathematical,
  physical and engineering sciences 463~(2088) (2007) 3251--3269.

\bibitem{huang2006sequential}
D.~Huang, T.~T. Allen, W.~I. Notz, R.~A. Miller, Sequential kriging
  optimization using multiple-fidelity evaluations, Structural and
  Multidisciplinary Optimization 32~(5) (2006) 369--382.

\bibitem{klein2015towards}
A.~Klein, S.~Bartels, S.~Falkner, P.~Hennig, F.~Hutter, Towards efficient
  bayesian optimization for big data, in: NIPS 2015 Bayesian Optimization
  Workshop, 2015.

\bibitem{zaytsev2016variable}
A.~Zaytsev, Variable fidelity regression using low fidelity function blackbox
  and sparsification, in: Symposium on Conformal and Probabilistic Prediction
  with Applications, Springer, 2016, pp. 147--164.

\bibitem{zaytsev2016reliable}
A.~Zaytsev, Reliable surrogate modeling of engineering data with more than two
  levels of fidelity, in: 2016 7th International Conference on Mechanical and
  Aerospace Engineering (ICMAE), IEEE, 2016, pp. 341--345.

\bibitem{kandasamy2016gaussian}
K.~Kandasamy, G.~Dasarathy, J.~B. Oliva, J.~Schneider, B.~P{\'o}czos, Gaussian
  process bandit optimisation with multi-fidelity evaluations, in: Advances in
  Neural Information Processing Systems, 2016, pp. 992--1000.

\bibitem{kandasamy2017multi}
K.~Kandasamy, G.~Dasarathy, J.~Schneider, B.~P{\'o}czos, Multi-fidelity
  bayesian optimisation with continuous approximations, in: Proceedings of the
  34th International Conference on Machine Learning-Volume 70, JMLR. org, 2017,
  pp. 1799--1808.

\bibitem{burnaev2011modeling}
E.~Burnaev, A.~Zaytsev, M.~Panov, P.~Prihodko, Y.~Yanovich, Modeling of
  nonstationary covariance function of gaussian process using decomposition in
  dictionary of nonlinear functions, Information Technologies and Systems--2011
  (2011) 2--7.

\bibitem{dolan2002benchmarking}
E.~D. Dolan, J.~J. Mor{\'e}, Benchmarking optimization software with
  performance profiles, Mathematical programming 91~(2) (2002) 201--213.

\bibitem{belyaev2016gtapprox}
M.~Belyaev, E.~Burnaev, E.~Kapushev, M.~Panov, P.~Prikhodko, D.~Vetrov,
  D.~Yarotsky, Gtapprox: Surrogate modeling for industrial design, Advances in
  Engineering Software 102 (2016) 29--39.

\end{thebibliography}

\end{document}